\documentclass[10pt,twocolumn,letterpaper]{article}

\usepackage{iccv}
\usepackage{times}
\usepackage{epsfig}
\usepackage{graphicx}
\usepackage{amsmath}
\usepackage{amssymb}
\usepackage{booktabs}
\usepackage{multirow}
\usepackage{subfig}

\usepackage[pagebackref=true,breaklinks=true,colorlinks,bookmarks=false]{hyperref}
\usepackage{makecell}
\DeclareSubrefFormat{parens}{#1(#2)}
\iccvfinalcopy 

\def\iccvPaperID{10937} 

\ificcvfinal\fi

\begin{document}

\title{Hierarchical Spatio-Temporal Representation Learning for Gait Recognition}

\author{Lei Wang\textsuperscript{1,2}, Bo Liu\textsuperscript{1,2}\thanks{Corresponding author: \href{mailto:boliu@hebau.edu.cn}{boliu@hebau.edu.cn}.}, Fangfang Liang\textsuperscript{1,2}~and Bincheng Wang\textsuperscript{1,2} \\
\textsuperscript{1} Hebei Agricultural University, \\ \textsuperscript{2} Hebei Key Laboratory of Agricultural Big Data
}

\maketitle
\ificcvfinal\thispagestyle{empty}\fi

\begin{abstract}
Gait recognition is a biometric technique that identifies individuals by their unique walking styles, which is suitable for unconstrained environments and has a wide range of applications. While current methods focus on exploiting body part-based representations, they often neglect the hierarchical dependencies between local motion patterns. In this paper, we propose a hierarchical spatio-temporal representation learning (HSTL) framework for extracting gait features from coarse to fine.  Our framework starts with a hierarchical clustering analysis to recover multi-level body structures from the whole body to local details. Next, an adaptive region-based motion extractor (ARME) is designed to learn region-independent motion features. The proposed HSTL then stacks multiple ARMEs in a top-down manner, with each ARME corresponding to a specific partition level of the hierarchy. An adaptive spatio-temporal pooling (ASTP) module is used to capture gait features at different levels of detail to perform hierarchical feature mapping. Finally, a frame-level temporal aggregation (FTA) module is employed to reduce redundant information in gait sequences through multi-scale temporal downsampling. Extensive experiments on CASIA-B, OUMVLP, GREW, and Gait3D datasets demonstrate that our method outperforms the state-of-the-art while maintaining a reasonable balance between model accuracy and complexity.


\end{abstract}

\section{Introduction}

Unlike other biometric technologies such as fingerprint, iris, and face, human gait can be captured at a distance without subject cooperation \cite{sepas2022deep}. By evaluating individual-specific walking patterns, gait recognition has been applied in a variety of fields, including criminal investigations \cite{muramatsu2013gait,makihara2020gait}, sports science \cite{kumar2021gait,echterhoff2018gait}, and 
smart transportation \cite{yang2016design}. However, the recognition can be challenging due to large variations in viewpoint \cite{liu2010view,kusakunniran2020review}, occlusion \cite{rida2019robust,uddin2019spatio}, and wearing \cite{yeoh2016clothing,yao2021collaborative}.
\begin{figure}[htbp]
  \centering
  \includegraphics[width=1.0\linewidth]{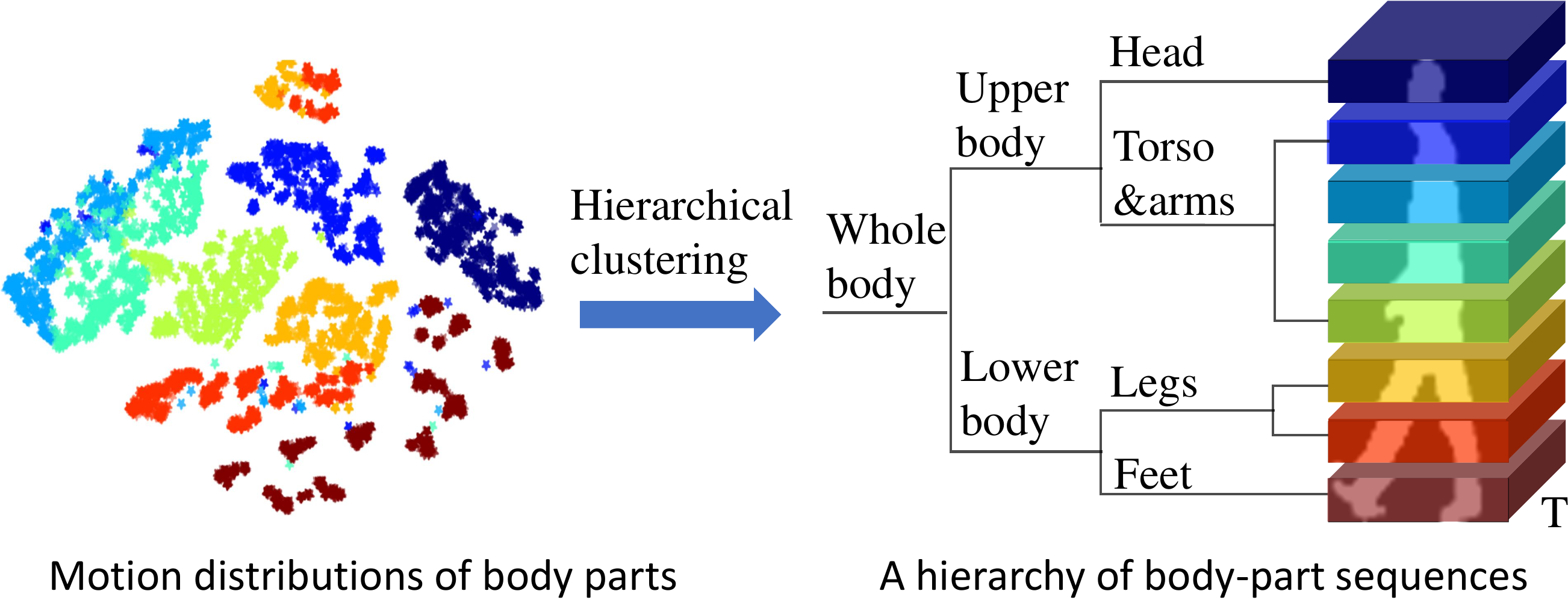}
\caption{The motivation for our approach. Left: the motion distributions of body parts. The same color indicates the same spatial location across multiple gait sequences in the CASIA-B dataset. Right: an example hierarchy of body-part sequences where $T$ denotes the temporal dimension of the sequence.}
\label{fig:moti}
\end{figure}

To address these issues, various approaches have been proposed for extracting gait features from silhouette sequences \cite{chao2019gaitset,lin2021gaitmask,huang20213d,huang2021context,zhang2021cross}, 3D human structures \cite{an2020performance,li2020end,teepe2021gaitgraph,zheng2022gait,li2021end}, or gait templates \cite{han2005individual,shiraga2016geinet,yu2017invariant}.  
 Silhouette-based gait recognition methods have gained increasing attention due to the ease of obtaining silhouettes from raw videos while preserving essential temporal information. The alignment of the input silhouette makes it possible for some methods to extract local body features by horizontally slicing the silhouette image \cite{zhang2019cross} or intermediate-layer features \cite{fan2020gaitpart,lin2021gait}. This partitioning strategy,  first introduced in person re-identification (ReID) \cite{sun2018beyond}, has been proven to be effective for gait recognition \cite{chao2019gaitset,fan2020gaitpart,hou2020gait,chai2022lagrange}.

However, the main limitation of the above part-based approaches is that they do not consider the hierarchical nature of local body movements \cite{bideau2018best}. For instance, within a gait cycle, the feet and lower body have distinct motion characteristics. Therefore, it is important to treat these body regions separately and investigate their part-whole relationships. Our motivation stems from the examination of body-part-specific motion clues. Specifically, each raw gait sequence in the CASIA-B \cite{yu2006framework} dataset is uniformly divided into eight part sequences along the body axis, so that each division roughly match a particular body part \footnote{Each part is pooled into a vector for visualization using t-SNE \cite{van2008visualizing}.}. The distributions of all body parts are shown in the left part of Fig.~\ref{fig:moti}. Observably, some parts, e.g., the head and feet, are easily separated owing to their large changes in walking kinematics. Whereas other parts, such as the thighs and calves, overlap due to the strong motion correlations between them. Further, to identify the relational structure among the part sequences, a hierarchical clustering analysis \cite{ester1996density} is performed. The results are shown in the right part of Fig.~\ref{fig:moti}, indicating that the semantic body regions can be captured in the higher clustering levels without precise localization of the body parts.

Following the above findings, we propose a novel hierarchical spatio-temporal representation learning (HSTL) framework for gait representation. The HSTL framework consists of multiple adaptive region-based motion extractor (ARME) modules, which are stacked to learn hierarchical motion patterns implied in a gait sequence (as shown in Fig.\ref{fig:moti}). In the ARME module, to account for inter-regional differences, non-shared 3D convolutions are used in correspondence with individual body regions. There regions are pre-identified by a hierarchical clustering process performed on fixed horizontal partitions, allowing each body region to cover one or more body parts.  Consequently, the deeper the ARME is, the more local features it tends to extract. Moreover, an adaptive spatio-temporal pooling (ASTP) module is proposed, which couples with an ARME module on the corresponding level to obtain hierarchical gait embeddings.


In addition, changes in gait speed or sampling frequency may result in several redundant frames in a gait sequence. Although several temporal fusion strategies have been proposed, they lose spatial information \cite{fan2020gaitpart,huang2021context} or lack adaptability \cite{lin2021gait,lin2021gaitmask}. To address this issue, we propose a frame-level temporal aggregation strategy (FTA). FTA fuses temporal features at multiple time steps, preserving significant motion information while compressing the sequence length. The main contributions of this paper are summarized as follows.



$\bullet$ We propose a hierarchical spatio-temporal representation learning (HSTL) framework for gait recognition. HSTL takes into account the dependencies of body regions in gait motions, ensuring simplicity and scalability of the architectural design.

$\bullet$ We introduce an adaptive region-based motion extractor (ARME) module to learn region-independent spatio-temporal representation for gait sequences, an adaptive spatio-temporal pooling (ASTP) module to perform hierarchical feature mapping, and a frame-level temporal aggregation (FTA) strategy to compress a gait sequence by removing redundant frames.

$\bullet$ Extensive experiments on the widely used gait datasets CASIA-B \cite{yu2006framework}, including OUMVLP \cite{NorikoTakemura2018MultiviewLP}, GREW \cite{zhu2021gait} and Gait3D \cite{zheng2022gait}, demonstrate that our method achieves state-of-the-art performance while offering a suitable trade-off between model accuracy and complexity.

\section{Related Work}
\label{sec:rela}
\subsection{Gait Recognition} Deep learning-based gait recognition methods can be broadly categorized into two categories: model-based and appearance-based. Model-based approaches extract structure and motion information from gait videos with the aid of pose estimation \cite{an2020performance,li2020end,teepe2021gaitgraph,li2021end,teepe2022towards,zheng2022gait}. Although these methods are robust to changes in viewpoint and appearance, they are sensitive to the accuracy of the pose parameters, making them incapable of handling low-resolution data. On the other hand, appearance-based approaches learn the feature
representation from raw videos \cite{zhang2019gait,song2019gaitnet,liang2022gaitedge}, or binary silhouette sequences \cite{han2005individual,shiraga2016geinet,yu2017invariant,chao2019gaitset,hou2020gait,fan2020gaitpart,hou2021set,lin2021gait,huang20213d,hou2022gait,chai2022lagrange,dou2022metagait}, which offer greater flexibility than model-based approaches. Our proposed method belongs to the family of appearance-based methods and uses silhouette sequences as inputs.
\begin{figure*}[htbp]
  \centering
  \includegraphics[width=1.00\linewidth]{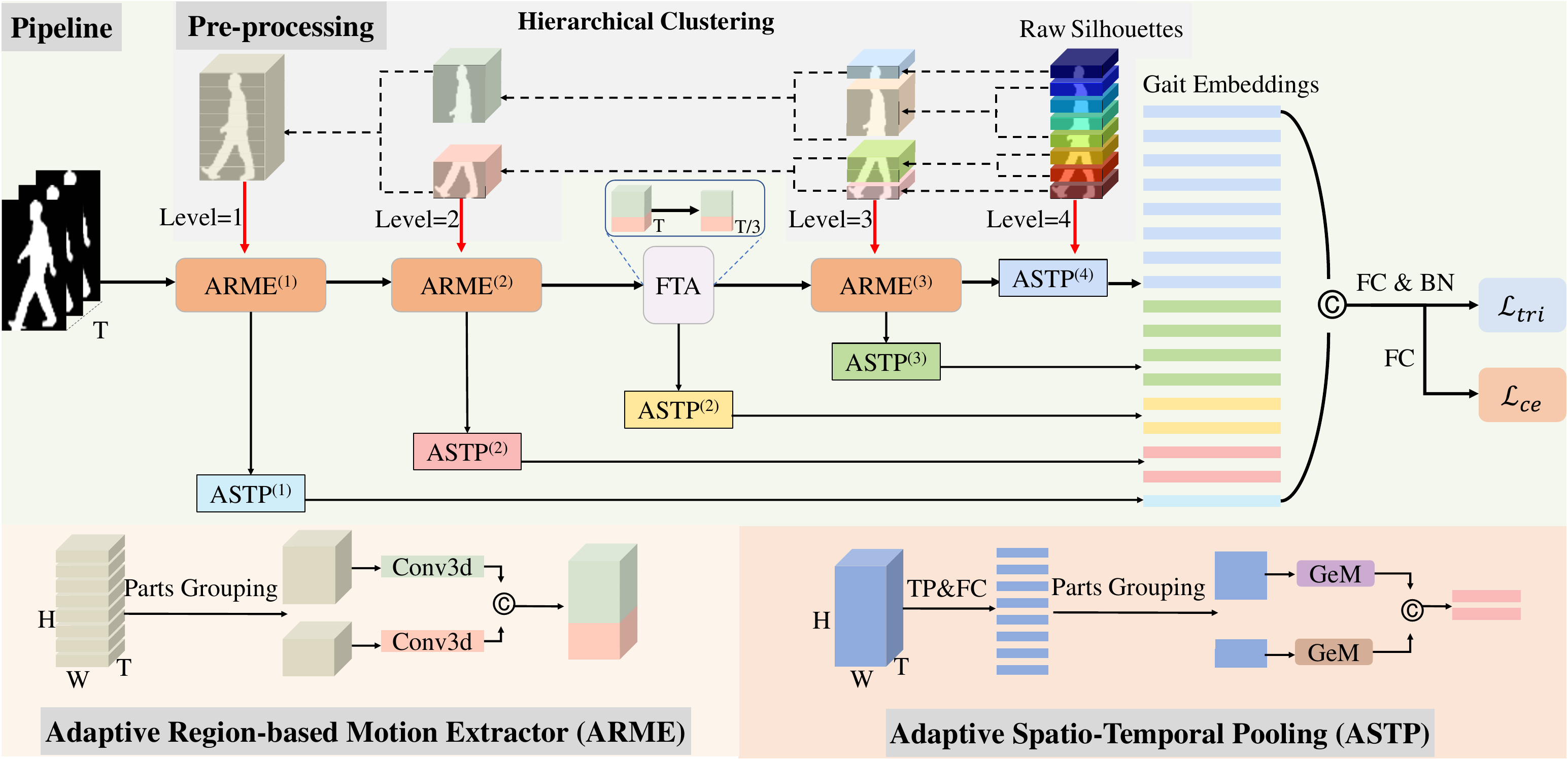}
\caption{The framework of HFSL. It mainly consists of three modules: ARME (adaptive region-based motion extractor), ASTP (adaptive spatio-temporal pooling), and FTA (frame-level temporal aggregation).  During pre-processing, a hierarchy of walking is obtained to guide the architectural design of HFSL. The framework uses multiple ARMEs to extract gait features from the entire body to individual regions. The ASTP module performs hierarchical feature mapping for the output of each level of ARME. The FTA module compresses local clips of each gait sequence to reduce the number of redundant frames. $T$, $H$ and $W$ denote dimensions of the feature maps. $\copyright$ represents the concatenation operation.}
\label{fig:method}
\end{figure*}


\subsection{Hierarchical Model} Hierarchical feature representation has been successfully applied to a wide range of vision tasks. Here, we provide a brief review of the hierarchical object ReID approaches related to gait recognition. 

In person ReID, some approaches  \cite{matsukawa2016hierarchical,zhang2022person,wang2021robust,tan2022dynamic,zhang2021coarse} hierarchically learn local descriptions and aggregate appearance features at different levels. For example, Matsukawa \etal \cite{matsukawa2016hierarchical} described an image patch via hierarchical Gaussian distribution. Zhang \etal \cite{zhang2021coarse} proposed a framework to learn coarse-grained and fine-grained features according to body structure. To solve the occlusion problem, Tan \etal \cite{tan2022dynamic} devised a hierarchical mask generator to learn from both occluded and holistic joint images. In vehicle ReID, some approaches \cite{wei2018coarse,shyam2021adversarially,li2022attribute} 
extract features from vehicle images in a hierarchical manner. For instance, Wei \etal \cite{wei2018coarse} proposed an RNN-based module for extracting latent cues from the model level to the vehicle level. Shyam \etal \cite{shyam2021adversarially} developed an attention-based hierarchical feature extractor. In addition, Li \etal \cite{li2022attribute} proposed a global structural embedding module for investigating hierarchical relationships between vehicle characteristics by incorporating attribute and state information.

For gait recognition, fusing features of multiple granularities can improve performance \cite{zhang2019cross,fan2020gaitpart,lin2021gait,lin2021gaitmask,huang20213d,chai2022lagrange}. In particular, CSTL \cite{huang2021context} proposed a temporal modeling network that integrates multi-scale temporal features adaptively. By combining part-level and sequence-level features, GaitPart \cite{fan2020gaitpart} obtained a part-independent spatio-temporal expression. Additionally, GaitGL \cite{lin2021gait} considered both full body-based and part-based information to achieve discriminative feature learning. Instead of equally dividing the feature maps, in 3D Local \cite{huang20213d},  a localization operation was developed to find 3D volumes of body parts in a sequence. However, most existing gait recognition methods do not sufficiently exploit the hierarchical dependencies among body parts during walking. In this paper, the proposed HSTL performs a coarse-to-fine hierarchical strategy that integrates multi-level motion patterns from gait sequences.
\subsection{Temporal Model}
Temporal cues play a crucial role in gait recognition due to the periodic changes in body shape. Previous methods treat a gait sequence as an unordered set, either compressing it into a single gait template during preprocessing \cite{shiraga2016geinet,li2020gait} or learning order-independent gait representations from silhouette sets \cite{chao2019gaitset,hou2020gait,hou2021set}. These methods assume that different subjects share similar global gait patterns, making ordering inputs unnecessary for gait assessment. However, ignoring the temporal nature of the gait sequence can result in missing discriminative local motion information. Recently, some approaches have achieved significant performance gains by explicitly modeling temporal information using LSTM \cite{zhang2019cross}, 1D convolution \cite{huang2021context}, and 3D convolution \cite{lin2020gait,huang20213d}. Nevertheless, these spatio-temporal operators also significantly increase computational costs. Although 
 some methods have been proposed to reduce video length by aggregating local clips \cite{lin2021gait,lin2021gaitmask,chai2022lagrange}, they lack adaptability to variations in pace. The main difference between our approach and others \cite{li2019selective,hou2021bicnet} is that we employ multi-scale temporal pooling at the frame level while considering variations in motion across body regions, leading to a more adaptable reduction of the gait sequence length.

\section{Proposed Method}
\label{sec:method}
In this section, we present the detailed description of HSTL, including the adaptive region-based motion extractor (ARME), the adaptive spatio-temporal pooling (ASTP), and the frame-level temporal aggregation (FTA).

\subsection{Framework Pipeline}
\label{sec:pipe}
The overview of our HSTL is presented in Fig.~\ref{fig:method}. Given a gait dataset $\mathcal{D}=\{{S_i}\}_{i=1}^{N} $ with $N$ gait sequences, where each sequence $S_i \in \mathbb{R}^{C \times T \times H \times W}$ is represented as a 4D tensor with $C$ channels, $T$ frames, and $H \times W$ pixels. During  the preprocessing stage, each gait sequence $S_i$  is divided horizontally and uniformly into $k$ part sequences, indexed from 1 to $k$. Then, a hierarchical clustering algorithm \cite{ester1996density} is applied to these part sequences to obtain a generic hierarchy of gait motions, which is denoted as $\mathcal{P}=\{\mathcal{P}^{(l)}\}_{l=1}^{L}$. Here, $L$ is the number of levels in the hierarchy and $\mathcal{P}^{(l)}$ is the set of partitions at level $l$. The partitions at level $l$ are defined as 
 $\mathcal{P}^{(l)}=\{P_1^{(l)}, P_2^{(l)}, \ldots, P_{K_{l}}^{(l)}\}$, where $P_j^{(l)}$ is the $j$-th subset of part indices and $K_l$ is the number of groups at level $l$. For instance, the top level $\mathcal{P}^{(1)}=\left\{\{1,2,\ldots, k-1,k\}\right\}$ means all the $k$ parts can be considered as a whole for the gait analysis. This hierarchy provides a structured  property of the gait motion patterns and can be utilized to guide gait feature extraction. To achieve this, our proposed HSTL employs three modules: ARME for extracting independent multi-granularity motion features, ASTP for generating vectorized gait embeddings, and FTA for reducing redundant information at the frame level. The HSTL stacks these three modules according to the division in $\mathcal{P}$, and the main branch of the HSTL for the input sequence $S_{in}$ can be formalized as:
\begin{equation}
\label{eq:pipeline}
    Y^{M}=\Gamma^{(L)} \circ \Psi^{(L-1)} \circ \cdots \circ \Omega^{(2)} \circ \Psi^{(2)} \circ \Psi^{(1)}(S_{in}),
\end{equation}
where $\Psi^{(l)}$, $\Gamma^{(l)}$, and $\Omega^{(l)}$ represent the ARME, ASTP and FTA modules at the $l$-th level of $\mathcal{P}$, respectively.  Since FTA uses inter-frame compression to reduce redundant information, it is employed only once at the $l_{\Omega}$-th level in $\mathcal{P}$ (e.g., $l_{\Omega}=2$ in Eq. (\ref{eq:pipeline})) to prevent excessive loss of information.

To obtain the hierarchical gait embeddings, the output $Y^{(l)}$ of each $\Psi^{(l)}$ at levels $l \in \{1, 2, \ldots, L-1\}$ and the output of $\Omega^{(l_{\Omega})}$ at level $l_{\Omega}$, denoted as $Y_{\Omega}^{(l_{\Omega})}$,  are fed into the corresponding $\Gamma^{(l)}$. The resulting outputs from these $L$ auxiliary branches are concatenated with the output of the main branch defined in Eq.~(\ref{eq:pipeline}), forming the final result, denoted as $Y$, which is given by:
\begin{align}
  Y =  \biggl[ & Y^{M}, \Gamma^{(L-1)}\left(Y^{(L-1)}\right),  \ldots , \nonumber \\ 
  & \Gamma^{(l_{\Omega})}\left(Y_{\Omega}^{(l_{\Omega})}\right), \Gamma^{(2)}\left(Y^{(2)}\right), \Gamma^{(1)}\left(Y^{(1)}\right) \biggl],
\end{align}
where $[,]$ denotes the concatenation operation.

Finally, $Y$ undergoes feature mapping through separate fully connected layers. The model is then trained using a combination of triplet loss $\mathcal{L}_{tri}$ and cross-entropy loss $\mathcal{L}_{ce}$, which is a commonly adopted practice in gait recognition \cite{hou2020gait,lin2021gait,huang2021context,huang20213d,dou2022metagait}.  Further details regarding the relevant modules are described in the following subsections.

\subsection{Adaptive Region-based Motion Extractor}
\label{sec:pyra}

The adaptive region-based motion extractor (ARME) aims to extract independent spatio-temporal patterns that are associated with different human body parts in a gait sequence. Unlike existing methods that uniformly slice gait images or sequences along the height axis \cite{zhang2019cross,fan2020gaitpart,lin2021gait,lin2021gaitmask}, ARME considers the inherent hierarchical relationships among different part sequences that are consistent with walking patterns. This allows ARME to effectively capture the unique walking kinematics of each part.

Given the hierarchical relation $\mathcal{P}$ introduced in Section \ref{sec:pipe}, ARME first divides the input sequence $X$ into $K_l$ regions based on the partition of the $l$-th level $\mathcal{P}^{(l)}$, resulting in the set of regions $\{X_{j}^{(l)}\}_{j=1}^{K_l}$, where $X_{j}^{(l)} \in \mathbb{R}^{C \times T \times H_j^{(l)} \times W}$. $H^{(l)}_j=\frac{\left|P_j^{(l)}\right|}{k}H$ is the height of the $j$-th region of the $l$-th level. Then the $l$-th level of ARME, $\Psi^{(l)}$, can be defined as follows:  
\begin{equation}
    Y^{(l)}_{\Psi}=\Psi^{(l)}(X^{(l)})=\left[f_{1}(X_{1}^{(l)}), f_{2}(X_{2}^{(l)}) ,\ldots, f_{K_l}(X_{K_l}^{(l)})  \right],
\end{equation}
where $f_{j}\left(\cdot\right)$ represents the independent 3D convolution operation applied to the $j$-th region. The output feature map $Y^{(l)}_{\Psi} \in \mathbb{R}^{C^{(l)} \times T \times H \times W}$ of level $l$ has $C^{(l)}$ channels. This module only modifies the number of channels, preserving the spatial and temporal resolutions of the input feature map.

\subsection{Adaptive Spatio-Temporal  Pooling}
\label{sec:astp}

It is a common procedure in gait recognition to obtain a compact and fixed-length feature representation by performing horizontal and uniform slicing of feature maps and strip-based pooling \cite{chao2019gaitset,fu2019horizontal,lin2020gait,lin2021gait,huang2021context}. However, a non-uniform division of the feature maps is better at capturing the gait motion characteristics. Thus, the adaptive spatio-temporal pooling (ASTP) is devised to construct hierarchical feature mapping (as shown in Fig.~\ref{fig:method}). Similar to the ARME module described in Section \ref{sec:pyra}, the hierarchy $\mathcal{P}$ enables us to obtain the $j$-th region of the $l$-th level, denoted as $X_j^{(l)}$. The corresponding ASTP, denoted as $\Gamma^{(l)}$, can be expressed as follow:
\begin{equation}
Y_{\Gamma,j}^{(l)}=\Gamma_j^{(l)}(X_j^{(l)})=\text{GeM}_j \circ \text{FC} \circ \text{Max}(X_j^{(l)}),
\end{equation}
where $\text{Max}(\cdot)$ represents a max pooling operation along the temporal dimension, $\text{FC}(\cdot)$ represents a fully connected layer, $\text{GeM}_j(\cdot)$ represents a generalized mean pooling (GeM) operation \cite{radenovic2018fine} for $j$-th region, and $\Gamma_j:\mathbb{R}^{C \times T \times H_j^{(l)} \times W} \mapsto \mathbb{R}^{C \times 1 \times H_j^{(l)} \times W} \mapsto \mathbb{R}^{C^{(l)} \times 1 \times H_j^{(l)} \times W} \mapsto \mathbb{R}^{C^{(l)} \times 1 \times 1 \times 1}$. Therefore, by concatenating outputs of $K_l$ regions, we can obtain $Y_{\Gamma}^{(l)}=\left[Y_{\Gamma,1}^{(l)},Y_{\Gamma,2}^{(l)},\ldots,Y_{\Gamma,K_l}^{(l)}\right]$, where $Y_{\Gamma}^{(l)} \in \mathbb{R}^{C_{}^{(l)}\times 1 \times  K_l \times 1}$ is the output of ASTP at level $l$.

\subsection{Frame-level Temporal Aggregation}
\label{sec:multi}

\begin{figure}[t]%
    \centering
    \includegraphics[width=1.0\linewidth]{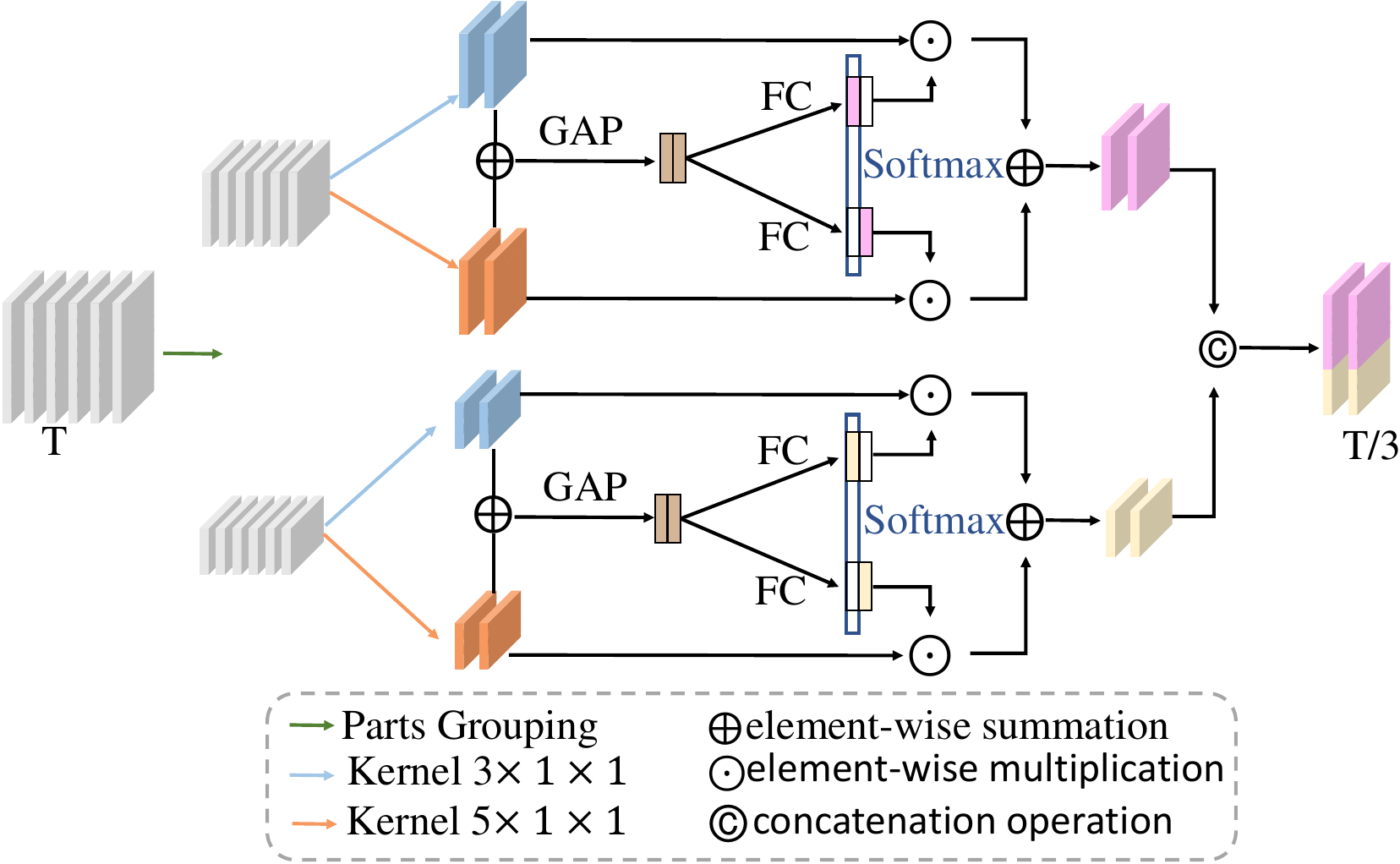}
    \caption{The detailed structure of the frame-level temporal aggregation (FTA). For simplicity, we omit the channel dimension $C$.
    }
    \label{fig:mma}
\end{figure}
A gait sequence may contain several redundant frames due to factors such as the acquisition frame rate and pace frequency. To reduce computational costs, some methods compress a gait sequence by aggregating its local clips \cite{lin2021gait,lin2021gaitmask}. In the proposed frame-level temporal aggregation (FTA) strategy, we consider both the gait structure and the multiscale temporal information. Given the $j$-th gait region at the $l$-th level, $X_{j}^{(l)}$, we first fuse the features of the two temporal scales using the following formula:
\begin{equation}
\begin{aligned}
\hat{U}_{j}^{(l)}&=U_{j,1}^{(l)}+U_{j,2}^{(l)}\\ &=\text{Max}_{3 \times 1 \times 1}^{3\times 1 \times 1}\left(X_{j}^{(l)}\right)
    +\text{Max}_{5 \times 1 \times 1}^{3\times 1 \times 1}\left(X_{j}^{(l)}\right),
\end{aligned} 
    \label{equ:poolfeature}
\end{equation}
where $\text{Max}_{3 \times 1 \times 1}^{3\times 1 \times 1}\left(\cdot\right)$ and $\text{Max}_{5 \times 1 \times 1}^{3\times 1 \times 1}\left(\cdot\right)$ denote max pooling operations with kernel sizes of $3\times 1 \times 1$ and $5\times 1 \times 1$ respectively, both with stride of $3 \times 1 \times 1$. $\hat{U}_{j}^{(l)}$, $U_{j,1}^{(l)}$ and  $U_{j,2}^{(l)}$ have the same size of $(C,\frac{T}{3},H_j^{(l)},W)$. 
The output of Eq.(\ref{equ:poolfeature}), $\hat{U}_{j}^{(l)}$, is the element-wise summation of the aggregation results of the two scales, $U_{j,1}^{(l)}$ and $U_{j,2}^{(l)}$, which reduces the temporal dimension of the input from  $T$ to $\frac{T}{3}$.

Then, the FTA model produces frame-level weights, which can be expressed as:
\begin{equation}
    \begin{gathered}
        Z_{j,1}^{(l)}=\text{FC}_{j,1}^{(l)}\left(\text{GAP}\left(\hat{U}_j^{(l)}\right)\right),\\
        Z_{j,2}^{(l)}=\text{FC}_{j,2}^{(l)}\left(\text{GAP}\left(\hat{U}_j^{(l)}\right)\right),
    \end{gathered}
        \label{equ:weight}
\end{equation}
where $\text{GAP}\left(\cdot\right)$ represents the global mean pooling along spatial dimension. 
$\text{FC}_{j,1}\left(\cdot\right)$ and $\text{FC}_{j,2}\left(\cdot\right)$ are two independent fully connected layers that generate the frame selection weighting tensors, $Z_{j,1}^{(l)}$ and $ Z_{j,2}^{(l)} \in \mathbb{R}^{C  \times \frac{T}{3} \times 1 \times 1}$, for $U_{j,1}^{(l)}$ and $U_{j,2}^{(l)}$, respectively. The weights are further normalized across the two scales, which can be written as follows:
\begin{equation}
  \label{equ:normalize}
       \mathcal{W}_{j,s,c,t}^{(l)}=\frac{e^{Z_{j,s,c,t}^{(l)}}}{e^{Z_{j,1,c,t}^{(l)}}+e^{Z_{j,2,c,t}^{(l)}}}   \quad s\in \{1,2\}, 
 \end{equation}      
where $\mathcal{W}_{j,s,c,t}^{(l)} \in \mathbb{R}^{1 \times 1 \times 1 \times 1}$ is the weight value of the $c$-th channel of the $t$-th frame.     
Combining Eq.~(\ref{equ:poolfeature}) and Eq.~(\ref{equ:normalize}), the $j$-th output region feature $Y_{\Omega,j}^{(l)} \in \mathbb{R}^{C^{(l)} \times \frac{T}{3} \times H_j^{(l)} \times W}$ for the $l$-th level of FTA can be obtained as follows:
\begin{equation} 
    Y_{\Omega,j}^{(l)}=\mathcal{W}_{j,1}^{(l)}\odot U_{j,1}^{(l)}+\mathcal{W}_{j,2}^{(l)}\odot U_{j,2}^{(l)},
\end{equation}
where $\mathcal{W}_{j,1}^{(l)}, \mathcal{W}_{j,2}^{(l)} \in \mathbb{R}^{C \times \frac{T}{3} \times 1 \times 1}$ are two weight tensors calculated using Eq.~(\ref{equ:poolfeature}), and $\odot$ represents element-wise multiplication operation. The FTA module outputs $Y^{(l)}_{\Omega}\in \mathbb{R}^{C \times \frac{T}{3} \times H \times W}$ by concatenating the $K_l$ gait regions of level $l$, where $Y^{(l)}_{\Omega}=\left[Y_{\Omega,1}^{(l)},Y_{\Omega,2}^{(l)},\ldots,Y_{\Omega,K_l}^{(l)}\right]$.

\begin{table*}[tbp]
  \centering
  \setlength{\abovecaptionskip}{0cm}  
\setlength{\belowcaptionskip}{-0.2cm} 
  \caption{Rank-1 accuracy (\%) on CASIA-B under all views and different conditions, excluding identical-view cases. Std denotes the performance sample standard deviation across 11 views.}
  \resizebox{0.95\linewidth}{!}{
    \begin{tabular}{c|c|ccccccccccc|c|c}
    \toprule
    \multicolumn{2}{c|}{Gallery NM \#1-4} & \multicolumn{11}{c|}{$0^{\circ}-180^{\circ}$}                                                         & \multirow{1}[4]{*}{Mean} & \multirow{1}[4]{*}{Std}\\
\cline{1-13}    \multicolumn{2}{c|}{Probe} & $0^{\circ}$    & $18^{\circ}$   & $36^{\circ}$   & $54^{\circ}$   & $72^{\circ}$   & $90^{\circ}$   & $108^{\circ}$  & $126^{\circ}$  & $144^{\circ}$  & $162^{\circ}$  & $180^{\circ}$  &  \\
    \hline
    \multirow{5}[10]{*}{NM \#5-6} & GaitSet \cite{chao2019gaitset} & 90.8  & 97.9  & 99.4  & 96.9  & 93.6  & 91.7  & 95.0    & 97.8  & 98.9  & 96.8  & 85.8  & 95.0 &3.5\\
          & GaitPart \cite{fan2020gaitpart} & 94.1  & 98.6  & 99.3  & 98.5  & 94.0    & 92.3  & 95.9  & 98.4  & 99.2  & 97.8  & 90.4  & 96.2 &3.1\\
         & 3D Local \cite{huang20213d} & 96.0    & \underline{99.0}    & \underline{99.5}  & \underline{98.9}  & 97.1  & 94.2  & 96.3  & 99.0    & 98.8  & 98.5  & 95.2  & 97.5&1.8 \\
          & CSTL \cite{huang2021context} & 97.2  & \underline{99.0}    & 99.2  & 98.1  & 96.2  & \underline{95.5}  & \underline{97.7}  & 98.7  & 99.2  & \underline{98.9}  & 96.5  & \underline{97.8} &1.3\\
          & GaitGL \cite{lin2021gait} & 96.0    & 98.3  & 99.0    & 97.9  & 96.9  & 95.4  & 97.0    & 98.9  & \underline{99.3}  & 98.8  & 94.0    & 97.4 &1.7\\
          & LagrangeGait \cite{chai2022lagrange} & 95.7  & 98.1  & 99.1  & 98.3  & 96.4  & 95.2  & 97.5  & 99.0    & \underline{99.3}  & \underline{98.9}  & 94.9  & 97.5 &1.6\\
         & MetaGait \cite{dou2022metagait} & \underline{97.3}  & \textbf{99.2}  & \underline{99.5}  & \textbf{99.1}  & \underline{97.2}  & \underline{95.5}  & 97.6  & \underline{99.1}    & \underline{99.3}  & \textbf{99.1}  & \underline{96.7}  & \textbf{98.1} & \underline{1.3} \\
\cline{2-15}          & \textbf{Ours} & \textbf{97.6}  & 98.0  & \textbf{99.6}    & 98.2    & \textbf{97.4}  & \textbf{96.5}  & \textbf{97.9}  & \textbf{99.3}  & \textbf{99.4}  & 98.4  & \textbf{97.0}  & \textbf{98.1} & \textbf{1.0}\\
    \hline
    \multirow{4}[12]{*}{BG \#1-2} & GaitSet \cite{chao2019gaitset} & 83.8  & 91.2  & 91.8  & 88.8  & 83.3  & 81.0    & 84.1  & 90.0    & 92.2  & 94.4  & 79.0    & 87.2 &4.9 \\
           & GaitPart \cite{fan2020gaitpart} & 89.1  & 94.8  & 96.7  & 95.1  & 88.3  & 84.9  & 89.0    & 93.5  & 96.1  & 93.8  & 85.8  & 91.5&4.2 \\
          & 3D Local \cite{huang20213d} & 92.9  & 95.9  & \textbf{97.8}  & 96.2  & 93.0    & 87.8  & 92.7  & 96.3  & 97.9  & 98.0    & 88.5  & 94.3&3.5 \\
           & CSTL \cite{huang2021context}  & 91.7  & 96.5  & 97.0    & 95.4  & 90.9  & 88.0    & 91.5  & 95.8  & 97.0    & 95.5  & 90.3  & 93.6 &3.0\\
           & GaitGL \cite{lin2021gait} & 92.6  & \underline{96.6}  & 96.8  & 95.5  & 93.5  & 89.3  & 92.2  & 96.5  & 98.2  & 96.9  & 91.5  & 94.5&2.8 \\
          & LagrangeGait \cite{chai2022lagrange} & \underline{94.2}  & 96.2  & \underline{96.8}  & 95.8  & 94.3  & 89.5  & 91.7  & 96.8  & 98.0    & 97.0    & 90.9  & 94.6&2.7 \\
          & MetaGait \cite{dou2022metagait} & 92.9  & \textbf{96.7}  & 97.1  & \underline{96.4}  & \underline{94.7}  & \underline{90.4}  & \underline{92.9}  & \textbf{97.2}  & \underline{98.5}    & \textbf{98.1}    & \underline{92.3}  & \underline{95.2} &\underline{2.6}\\
\cline{2-15}          & \textbf{Ours} & \textbf{95.0}  & 96.5  & \underline{97.3}  & \textbf{96.6}  & \textbf{95.3}  & \textbf{93.3}  & \textbf{94.6}  & \underline{96.8}  & \textbf{98.6}  & \underline{97.7}  & \textbf{92.9}    & \textbf{95.9}&\textbf{1.7} \\
    \hline
    \multirow{4}[12]{*}{CL \#1-2} & GaitSet \cite{chao2019gaitset} & 61.4  & 75.4  & 80.7  & 77.3  & 72.1  & 70.1  & 71.5  & 73.5  & 73.5  & 68.4  & 50.0    & 70.4&8.0 \\
         & GaitPart \cite{fan2020gaitpart} & 70.7  & 85.5  & 86.9  & 83.3  & 77.1  & 72.5  & 76.9  & 82.2  & 83.8  & 80.2  & 66.5  & 78.7 &6.6\\
          & 3D Local \cite{huang20213d} & 78.2  & 90.2  & 92.0    & 87.1  & 83.0    & 76.8  & 83.1  & 86.6  & 86.8  & 84.1  & 70.9  & 83.7 &6.2\\
          & CSTL \cite{huang2021context}  & 78.1  & 89.4  & 91.6  & 86.6  & 82.1  & 79.9  & 81.8  & 86.3  & 88.7  & 86.6  & 75.3  & 84.2 &\underline{4.9}\\
          & GaitGL \cite{lin2021gait} & 76.6  & 90.0    & 90.3  & 87.1  & 84.5  & 79.0    & 84.1  & 87.0    & 87.3  & 84.4  & 69.5  & 83.6&6.3 \\
          & LagrangeGait \cite{chai2022lagrange} & 77.4  & 90.6  & \underline{93.2}  & \underline{90.2}  & 84.7  & 80.3  & \underline{85.2}  & 87.7  & 89.3  & 86.6  & 71.0    & 85.1 &6.3\\
          & MetaGait \cite{dou2022metagait} & \underline{80.0}  & \underline{91.8}  & 93.0  & 87.8  & \underline{86.5}  & \underline{82.9}  & \underline{85.2}  & \underline{90.0}  & \underline{90.8}  & \underline{89.3}  & \underline{78.4}    & \underline{86.9} &\textbf{4.6}\\
\cline{2-15}          & \textbf{Ours} & \textbf{82.4}  & \textbf{94.2}  & \textbf{95.0}  & \textbf{91.7}  & \textbf{88.2}  & \textbf{83.3}  & \textbf{88.0}  & \textbf{92.3}  & \textbf{93.1}  & \textbf{91.0}  & \textbf{78.5}  & \textbf{88.9} &5.1\\
    \bottomrule
    \end{tabular}%
    }
  \label{tab:sota-cb}%
\end{table*}%
\section{Experiments}
\label{sec:exper}
\subsection{Datasets and Evaluation Protocols}
\label{sec:dataset}
\noindent\textbf{CASIA-B.} The CASIA-B \cite{yu2006framework} dataset is a widely used benchmark for gait recognition. It contains video sequences of 124 subjects with 11 different views and three walking conditions (normal walking (NM), walking with a bag (BG), and walking with a coat (CL)). Our study follows the protocol outlined in previous works \cite{chao2019gaitset,fan2020gaitpart,lin2021gait,lin2021gaitmask,huang2021context,huang20213d}. The first 74 subjects are used for training, and the remaining 50 subjects are used for testing. During testing, the first four sequences under NM (NM\#01-04) are regarded as the gallery set, and the rest  (NM\#05-06, BG\#01-02, CL\#01-02) are regarded as the probe set.

\noindent\textbf{OUMVLP.} The OUMVLP \cite{NorikoTakemura2018MultiviewLP} is one of the largest gait datasets, containing silhouette sequences of 10,307 subjects. Each subject has a single normal walking condition (NM) with 14 views. According to the protocol provided by the dataset, the first 5,153 subjects are used for training while the remaining 5,154 subjects are used for testing. During the testing phase, the sequences of NM\#01 are assigned to the gallery set, and the sequences of NM\#02 are considered as the probe set.

\noindent\textbf{GREW.} GREW \cite{zhu2021gait} is the first large-scale dataset for gait recognition in the wild, consisting of 128,671 sequences from 26,345 individuals captured by 882 cameras. It includes four data types: four data types: silhouettes, optical flow, 2D pose, and 3D pose. The dataset is divided into a training set with 20,000 subjects and 102,887 sequences, and a testing set with 6,000 subjects and 24,000 sequences. In the testing phase, each subject has two sequences for the gallery set and two for the probe set. The GREW dataset also includes a distractor set, which contains 233,857 unlabeled sequences.

\noindent\textbf{Gait3D.} The Gait3D dataset \cite{zheng2022gait} is a newly proposed dataset for gait recognition in uncontrolled indoor environments, particularly in large supermarkets. It contains 25,309 sequences of 4,000 subjects extracted from 39 cameras, with 18,940 sequences from 3,000 subjects for training and 6,369 sequences from 1,000 subjects for testing. The dataset mainly includes four data types: silhouettes, 2D pose, 3D pose, and 3D mesh. During testing, one sequence per subject is used as the probe set, while the remaining sequences are used as the gallery set.  To evaluate the model's performance, we use accuracy as well as mean average precision (mAP) and mean inverse negative penalty (mINP) \cite{ye2021deep}, which consider multiple instances and hard sample recall.

\subsection{Implementation Details}
\label{sec:impl}

\noindent \textbf{Training details.} 1) In our implementation, the margin $m$ of the triplet loss is set to 0.2, and the parameter $p$ of the GeM function used in the ASTP module is initialized to 6.5. In the hierarchy of gait motion, the number of partitions $k$ at the bottom level is set to 8. 2) The batch size is set to (8,8) for CASIA-B, (32,8) for OUMVLP, (32,4) for GREW, and (32,4) for Gait3D. 3) We use gait silhouettes as the input modality. During the training phase, we sample 30 frames following the strategy proposed in \cite{fan2020gaitpart}, while during testing, all frames are fed into the model. Moreover, we align each frame following the strategy presented in \cite{NorikoTakemura2018MultiviewLP}, and the input image size is cropped to $64 \times 44$ for all datasets. 4) For CASIA-B, the optimizer used is Adam with a weight decay of 5$e$-4. The model is trained for 100K iterations with an initial learning rate (LR) of 1$e$-5, and the LR is multiplied by 0.1 at 70K iterations. For OUMVLP, GREW, and Gait3D, the model is trained for 250K, 250K, and 210K iterations, respectively, with an initial LR of 0.1. The LR is multiplied by 0.1 at 150K and 200K iterations. The optimizer used for these datasets is SGD with a weight decay of 5$e$-4.

\begin{table}[bt]
  \centering
  \setlength{\abovecaptionskip}{0cm}  
\setlength{\belowcaptionskip}{-0.2cm} 
  \caption{The detailed architecture of the proposed HSTL on CASIA-B. The first column denotes the levels of the gait hierarchy and $K_l$ is the number of groups at level $l$. $C_{in}$ and $C_{out}$ represent the input channel and output channel of each layer respectively. The body parts are indexed in spatial order from top to bottom, numbered 1 to 8.}
  \resizebox{\linewidth}{!}{
    \begin{tabular}{c|ccccc|c|c}
\toprule    \multicolumn{1}{l|}{Level} & \multicolumn{1}{c|}{Block} & \multicolumn{1}{c|}{Layer} & \multicolumn{1}{c|}{$C_{in}$} & \multicolumn{1}{c|}{$C_{out}$} & Kernel & $K_l$  & \multicolumn{1}{c}{Parts Grouping} \\
\hline    \multicolumn{1}{c|}{\multirow{2}[1]{*}{1}}     & \multicolumn{1}{c|}{ARME} & \multicolumn{1}{c|}{Conv3d} & \multicolumn{1}{c|}{1} & \multicolumn{1}{c|}{32} & (3,3,3) & \multirow{2}[1]{*}{1} & \multirow{2}[1]{*}{$\{\{1,2,3,4,5,6,7,8\}\}$} \\
\cline{2-6}          & \multicolumn{5}{c|}{ASTP}             &       &  \\
\cline{1-8}      \multicolumn{1}{c|}{\multirow{3}[1]{*}{2}}    & \multicolumn{1}{c|}{\multirow{2}[1]{*}{ARME}} & \multicolumn{1}{c|}{Conv3d} & \multicolumn{1}{c|}{32} & \multicolumn{1}{c|}{32} & (3,3,3) & \multirow{3}[1]{*}{2} & \multirow{3}[1]{*}{$\{\{1,2,3,4,5\}$,$\{6,7,8\}\}$} \\
\cline{3-6}          & \multicolumn{1}{c|}{} & \multicolumn{1}{c|}{Conv3d} & \multicolumn{1}{c|}{32} & \multicolumn{1}{c|}{64} & (3,3,3) &       &  \\
\cline{2-6}         & \multicolumn{5}{c|}{ASTP}             &       &  \\
\cline{1-8}      \multicolumn{1}{c|}{\multirow{3}[1]{*}{2}}    & \multicolumn{1}{c|}{\multirow{2}[1]{*}{FTA}} & \multicolumn{1}{c|}{\multirow{2}[1]{*}{MaxPool}} & \multicolumn{1}{c|}{\multirow{2}[1]{*}{64}} & \multicolumn{1}{c|}{\multirow{2}[1]{*}{64}} & (3,1,1) & \multirow{3}[1]{*}{2} & \multirow{3}[1]{*}{$\{\{1,2,3,4,5\}$,$\{6,7,8\}\}$} \\
\cline{6-6}          & \multicolumn{1}{c|}{} & \multicolumn{1}{c|}{} & \multicolumn{1}{c|}{} & \multicolumn{1}{c|}{} & (5,1,1) &       &  \\
\cline{2-6}          & \multicolumn{5}{c|}{ASTP}             &       &  \\
\cline{1-8}       \multicolumn{1}{c|}{\multirow{3}[1]{*}{3}}   & \multicolumn{1}{c|}{\multirow{2}[1]{*}{ARME}} & \multicolumn{1}{c|}{Conv3d} & \multicolumn{1}{c|}{64} & \multicolumn{1}{c|}{128} & (3,3,3) & \multirow{3}[1]{*}{4} & \multirow{3}[1]{*}{$\{\{1\}$,$\{2,3,4,5\}$,$\{6,7\}$,$\{8\}\}$} \\
\cline{3-6}          & \multicolumn{1}{c|}{} & \multicolumn{1}{c|}{Conv3d} & \multicolumn{1}{c|}{128} & \multicolumn{1}{c|}{128} & (3,3,3) &       &  \\
\cline{2-6}          & \multicolumn{5}{c|}{ASTP}             &       &  \\
\cline{1-8}      4    & \multicolumn{5}{c|}{ASTP}             & 8    & \makecell[c]{$\{\{1\},\{2\},\{3\},\{4\},$\\$\{5\},\{6\},\{7\},\{8\}\}$} \\
\bottomrule    \end{tabular}%
}
  \label{tab:pcb-str}%
\end{table}%

\vspace{-0.13cm}
\noindent \textbf{Architecture details.} Table ~\ref{tab:pcb-str} presents the detailed architecture of the model used for the CASIA-B dataset. To handle datasets with more subjects, such as OUMVLP, GREW, and Gait3D, we incorporate the label smoothing operation into the cross-entropy loss function, and deepen the network by adding an extra ARME module to the third level of the hierarchy. The output channels for the four ARMEs are set to 64, 64, 128 and 256, respectively.
Additionally, we include a layer of spatial downsampling after the first ARME to improve the training efficiency.




\begin{table*}[tbp]
  \centering
  \setlength{\abovecaptionskip}{0cm}  
\setlength{\belowcaptionskip}{-0.2cm} 
  \caption{Rank-1 accuracy (\%) on OUMVLP under all views, excluding identical-view cases. Std denotes the performance sample standard deviation across 14 views.}
  \resizebox{0.95\linewidth}{!}{
    \begin{tabular}{c|cccccccccccccc|c|c}
    \toprule
    \multirow{2}[1]{*}{Method} & \multicolumn{14}{c|}{Probe View}                                                                              & \multirow{2}[1]{*}{Mean}&\multirow{2}[1]{*}{Std} \\
\cline{2-15}          & $0^{\circ}$    & $15^{\circ}$   & $30^{\circ}$   & $45^{\circ}$   &$60^{\circ}$   & $75^{\circ}$   & $90^{\circ}$   & $180^{\circ}$  & $195^{\circ}$  & $210^{\circ}$  & $225^{\circ}$  & $240^{\circ}$  & $255^{\circ}$  & $270^{\circ}$  &  \\
    \hline
    GaitSet \cite{chao2019gaitset} & 79.3  & 87.9  & 90.0    & 90.1  & 88.0    & 88.7  & 87.7  & 81.8  & 86.5  & 89.0    & 89.2  & 87.2  & 87.6  & 86.2  & 87.1 &4.0 \\
    
    GaitPart \cite{fan2020gaitpart} & 82.6  & 88.9  & 90.8  & 91.0    & 89.7  & 89.9  & 89.5  & 85.2  & 88.1  & 90.0    & 90.1  & 89.0    & 89.1  & 88.2  & 88.7 &2.3\\
    
    GLN \cite{hou2020gait}   & 83.8  & 90.0    & 91.0    & 91.2  & 90.3  & 90.0    & 89.4  & 85.3  & 89.1  & 90.5  & 90.6  & 89.6  & 89.3  & 88.5  & 89.2 & 2.1\\
    
    CSTL \cite{huang2021context}  & 87.1  & 91.0    & 91.5  & 91.8  & 90.6  & 90.8  & 90.6  & \underline{89.4}  & 90.2  & 90.5  & 90.7  & 89.8  & 90.0    & 89.4  & 90.2 & \underline{1.1}\\
    
    GaitGL \cite{lin2021gait} & 84.9  & 90.2  & 91.1  & 91.5  & 91.1  & 90.8  & 90.3  & 88.5  & 88.6  & 90.3  & 90.4  & 89.6  & 89.5  & 88.8  & 89.7&1.7 \\
    
    3D Local \cite{huang20213d} & 86.1  & 91.2  & 92.6  & 92.9  & 92.2  & 91.3  & 91.1  & 86.9  & 90.8  & \textbf{92.2}  & 92.3  & 91.3  & 91.1  & 90.2  & 90.9&2.0 \\
    
    LagrangeGait \cite{chai2022lagrange} & 85.9  & 90.6  & 91.3  & 91.5  & 91.2  & 91.0    & 90.6  & 88.9  & 89.2  & 90.5  & 90.6  & 89.9  & 89.8  & 89.2  & 90.0 & 1.4\\
    
    MetaGait \cite{dou2022metagait} & \underline{88.2}  & \underline{92.3}  & \textbf{93.0}  & \textbf{93.5}  & \textbf{93.1}  & \textbf{92.7}    & \textbf{92.6}  & 89.3  & \underline{91.2}  & 92.0  & \textbf{92.6}  & \textbf{92.3}  & \textbf{91.9}  & \underline{91.1}  & \underline{91.9} & 1.4 \\
    \hline
    \textbf{Ours}  &   \textbf{91.4}    &  \textbf{92.9}     &   \underline{92.7}    &  \underline{93.0}     &  \underline{92.9}     &   \underline{92.5}    &  \underline{92.5}     &  \textbf{92.7}     &   \textbf{92.3}    &   \underline{92.1}    &  \underline{92.3}     &   \underline{92.2}    &  \underline{91.8}     &   \textbf{91.8}    & \textbf{92.4} & \textbf{0.5}\\
    \bottomrule
    \end{tabular}%
  \label{tab:sota-ou}%
  }
\end{table*}%

\begin{table}[htbp]
  \centering
  \setlength{\abovecaptionskip}{0cm}  
\setlength{\belowcaptionskip}{-0.2cm} 
  \caption{Rank-1 accuracy (\%), Rank-5 accuracy (\%), Rank-10 accuracy (\%), Rank-20 accuracy (\%) on GREW.}
  \resizebox{0.8\linewidth}{!}{
    \begin{tabular}{c|cccc}
    \toprule
    Methods & Rank-1 & Rank-5 & Rank-10 & Rank-20 \\
    \hline
    GaitSet \cite{chao2019gaitset} & 46.28 & 63.58 & 70.26 & 76.82 \\
    GaitPart \cite{fan2020gaitpart} & 44.01 & 60.68 & 67.25 & 73.47 \\
    GaitGL \cite{lin2021gait} & 47.28 & 63.56 & 69.32 & 74.18 \\
    MTSGait \cite{zheng2022mts} & \underline{55.32} & \underline{71.28} & \underline{76.85} & \underline{81.55} \\
    \hline
    \textbf{Ours} & \textbf{62.72} & \textbf{76.57} & \textbf{81.32} & \textbf{85.24} \\
    \bottomrule
    \end{tabular}%
    }
  \label{tab:grew}%
\end{table}%

\begin{table}[htbp]
  \centering
  \setlength{\abovecaptionskip}{0cm}  
\setlength{\belowcaptionskip}{-0.2cm} 
  \caption{Rank-1 accuracy (\%), Rank-5 accuracy (\%), mAP (\%) and mINP on Gait3D.}
  \resizebox{0.8\linewidth}{!}{
    \begin{tabular}{c|cccc}
    \toprule
    Methods & Rank-1 & Rank-5 & mAP   & mINP \\
    \hline
    GaitSet \cite{chao2019gaitset} & 36.70  & 58.30  & 30.01  & 17.30  \\
    GaitPart \cite{fan2020gaitpart} & 28.20  & 47.60  & 21.58  & 12.36  \\
    GLN \cite{hou2020gait}  & 31.40  & 52.90  & 24.74  & 13.58  \\
    GaitGL \cite{lin2021gait} & 29.70  & 48.50  & 22.29  & 13.26  \\
    CSTL \cite{huang2021context} & 11.70  & 19.20  & 5.59  & 2.59  \\
    SMPLGait \cite{zheng2022gait} & 46.30  & 64.50  & 37.16  & \underline{22.23}  \\
    MTSGait \cite{zheng2022mts} & \underline{48.70}  & \underline{67.10}  & \underline{37.63}  & 21.92  \\
    \hline
    \textbf{Ours} & \textbf{61.30} & \textbf{76.30} & \textbf{55.48} & \textbf{34.77} \\
    \bottomrule
    \end{tabular}%
    }
  \label{tab:res-gait3d}%
\end{table}%

\subsection{Comparison with State-of-the-Art Methods}
\label{sec:compar}
\noindent\textbf{Evaluation on CASIA-B.} Table~\ref{tab:sota-cb} compares the performance of the proposed HSTL with seven state-of-the-art methods on the CASIA-B dataset. It can be seen that the proposed method obtains the best results in all three walking conditions while maintaining considerable stability across different views. The experimental results reveal that 1) the mean accuracy of the proposed method for the BG and CL walking conditions is 95.9\% and 88.9\%, respectively, which are higher by 0.7\% and 2.0\% than the second-best method (MetaGait \cite{dou2022metagait}),  demonstrating the advantage of our method in cross-view gait recognition. 2) 
The decrease in accuracy from NM to CL is 9.2\% for our method, compared to 12.4\% for 3D Local \cite{huang20213d}.
 This indicates that the ARME module is more adaptable to various walking conditions.  In addition, our method achieves an accuracy of 88.9\% in the challenging CL condition, which is 5.2\% higher than 3D Local. This may be due to the complex clothing conditions that affect the part localization accuracy in 3D Local. 
 3) Both CSTL \cite{huang2021context} and our method extracts multi-scale temporal features but in different ways. CSTL first extracts spatial information and then fuses three scales of motion features. In contrast, we propose an FTA module to aggregate spatio-temporal information from multiple body regions. Therefore, the average rank-1 accuracy of our method is higher or equivalent to that of CSTL in all views. This suggests that FTA can be more adaptive to spatio-temporal changes.

\begin{figure}[tb]
\centering
    \subfloat[Average rank-1 accuracy]{
        \includegraphics[width=0.61\linewidth]{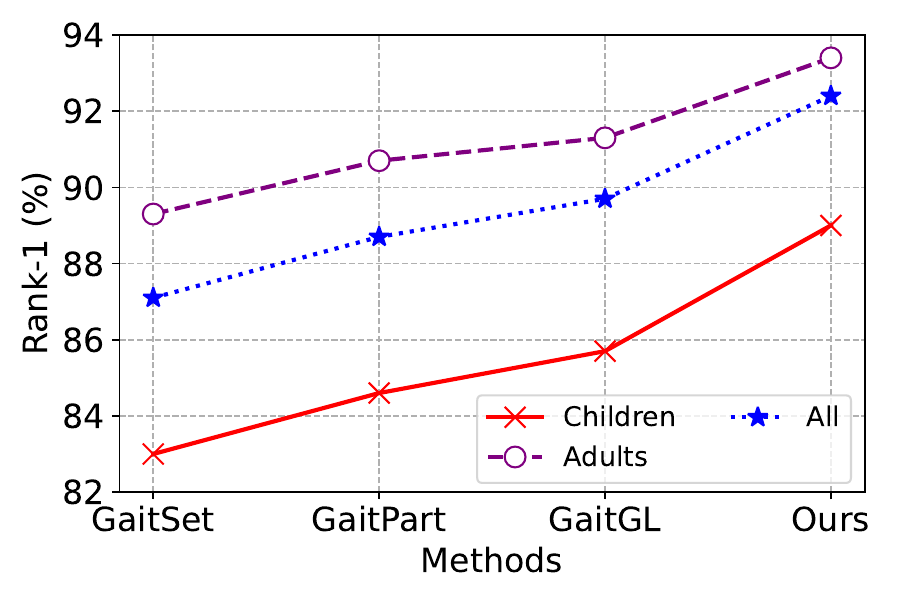}
        \label{fig:age}
        }
    \subfloat[Age distribution]{
        \includegraphics[width=0.35\linewidth]{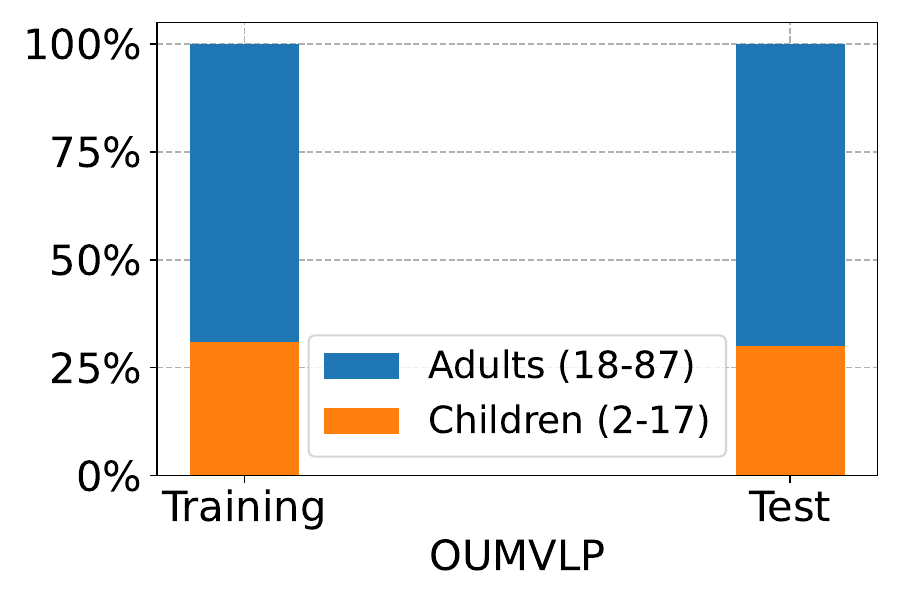}
        \label{fig:prob}
        }
    \caption{Cross-age comparison results for the OUMVLP dataset. (a) Average rank-1 accuracy in adults vs children. (b) Distribution of two age groups.}
    \label{fig:cvpr-ou}
\end{figure}

\noindent\textbf{Evaluation on OUMVLP.} To verify the model's generalizability, we conduct experiments on the large-scale OUMVLP dataset.  As shown in Table~\ref{tab:sota-ou}, our approach achieves competitive results in most views. In particular, the proposed method outperforms the second-best method (MetaGait) by an average of 2.5\% under three extreme views, i.e., 0$^{\circ}$, 180$^{\circ}$, and 270$^{\circ}$, resulting in the best mean performance and cross-view stability. In addition, the OUMVLP dataset also provides annotations for the ages of the subjects. To evaluate the impact of age differences on recognition performance, we divided all subjects into two groups: adults (18-87 years old) and children (2-17 years old), as shown in Fig.~\ref{fig:cvpr-ou}\subref{fig:prob}.  Fig.~\ref{fig:cvpr-ou}\subref{fig:age} presents the recognition accuracy based on age. It can be observed that, as adult sequences make up about 70\% of the total sequences, all the compared methods show a bias toward the recognition results of adults. However, compared to other methods, our model effectively improves the accuracy of gait recognition for children, demonstrating the effectiveness of our hierarchical gait representation across ages.

\noindent\textbf{Evaluation on GREW and Gait3D.}  Gait3D and GREW are two recently introduced datasets that contain challenging conditions, such as misalignment of the human body and partial occlusion. Tables ~\ref{tab:grew} and ~\ref{tab:res-gait3d} show the results of the comparison between the proposed method and the state-of-the-art methods. Our method shows superior performance in all metrics, indicating its ability to effectively model gait characteristics in realistic scenarios.

\begin{figure}[bt]
  \centering
  \includegraphics[width=0.87 \linewidth]{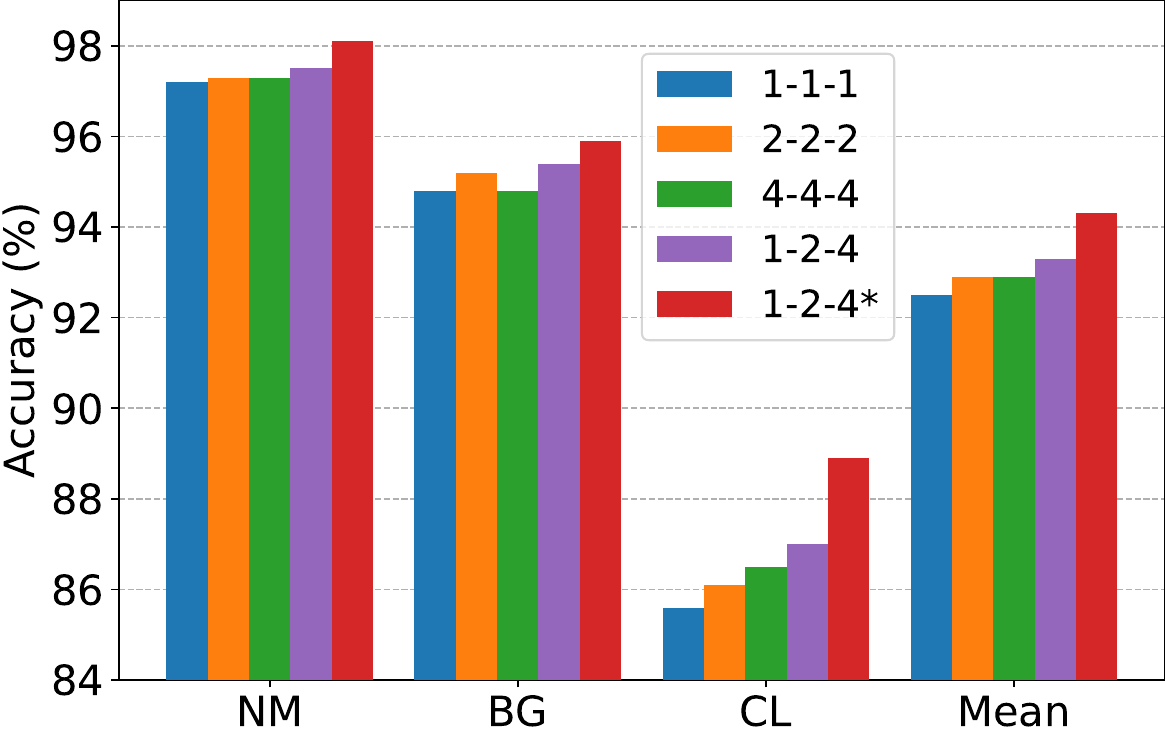}
\caption{Ablation study on the effectiveness of hierarchical feature extraction on the CASIA-B dataset (best viewed in color).}
\label{fig:hlb-ab}
\end{figure}
\subsection{Ablation Study}
\label{sec:albation}

\noindent\textbf{Effectiveness of hierarchical feature extraction.} To verify the effectiveness of our hierarchical gait partitioning, we conduct experiment with various grouping strategies. As shown in Table \ref{tab:pcb-str}, this comparison only considers the different settings of the first three layers since a uniform partition is used at layer 4. Specifically, 1-1-1 indicates that the first three levels have the same number of groupings, and they are divided uniformly. 1-2-4* refers to the non-uniform division used in the proposed model. As shown in Fig.~\ref{fig:hlb-ab}, the hierarchical feature extraction setting, e.g., 1-2-4, outperforms the other non-hierarchical approaches. A further mean performance improvement of 1.0\% is achieved when the motion relationships between different body regions, i.e., 1-2-4*, are considered.

 \noindent\textbf{Effectiveness of ARME, ASTP, and FTA.} The results of the ablation experiments for ARME, ASTP, and FTA are shown in Tab.~\ref{tab:impact} The results indicate that the ARME module significantly contributes to the improvement of the recognition accuracy, with an average improvement of 1.1\% compared to the baseline model (non-grouping version of ARME). The mean accuracy is further improved by 1.3\% when the ASTP module is integrated and multi-scale fusion is utilized in FTA. These results demonstrate the effectiveness and complementarity of the three modules in the proposed gait recognition framework.

\subsection{Trade-off between accuracy and efficiency}
\label{sec:trade-off}

\begin{figure}[tb]
  \centering
  \includegraphics[width=0.91\linewidth]{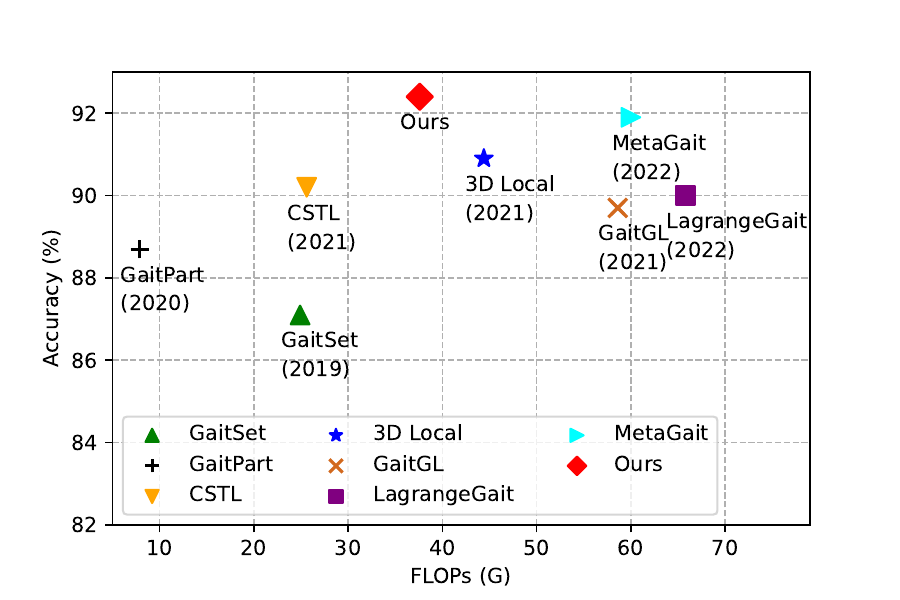}
\caption{The trade-off between accuracy and FLOPs of our method and other comparison methods on the OUMVLP dataset.}
\label{fig:trade-off4}
\end{figure}

 \begin{table}[tb]
  \centering
    \setlength{\abovecaptionskip}{0cm}  
\setlength{\belowcaptionskip}{-0.2cm} 
  \caption{Ablation study on the effectiveness of ARME, ASTP, and FTA modules in terms of average rank-1 accuracy on the CASIA-B dataset.}
  \resizebox{\linewidth}{!}{
    \begin{tabular}{c|cc|cc|ccc|c}
    \toprule
    \multirow{2}[1]{*}{Setting} & \multirow{2}[1]{*}{ARME} & \multirow{2}[1]{*}{ASTP} & \multicolumn{2}{|c|}{FTA} & \multirow{2}[1]{*}{NM} & \multirow{2}[1]{*}{BG} & \multirow{2}[1]{*}{CL} & \multirow{2}[1]{*} {Mean} \\
\cline{4-5}          &       &       &  \multicolumn{1}{|c|}3     & 5     &       &       &       &  \\
    \hline
     a     &      &       &       &       & 97.0  & 94.5  & 84.2  & 91.9 \\
    b     & \checkmark     &       &       &       & 97.8  & 95.3  & 85.9  & 93.0 \\
    
    c     & \checkmark     & \checkmark     &  &   & 97.8 & 95.4 & 87.0 & 93.4 \\
    d     & \checkmark     & \checkmark     & \checkmark     &       & 97.7  & 95.3  & 87.5  & 93.5 \\
    e     & \checkmark     & \checkmark     &       & \checkmark     & 97.7  & 95.2  & 87.2  & 93.4 \\
    f     & \checkmark     &       & \checkmark     & \checkmark     & 97.8  & 95.4  & 87.5  & 93.6 \\
    g     & \checkmark     & \checkmark     & \checkmark     & \checkmark     & \textbf{98.1}  & \textbf{95.9}  & \textbf{88.9}  & \textbf{94.3} \\
    
    
    \bottomrule
    \end{tabular}%
    }
  \label{tab:impact}%
\end{table}%
In this subsection, we evaluate the relationship between accuracy and efficiency for each compared method. As shown in Fig.~\ref{fig:trade-off4}, the 3D convolution-based approaches, such as GaitGL \cite{lin2021gait}, 3D Local \cite{huang20213d}, LagrangeGait \cite{chai2022lagrange} and MetaGait \cite{dou2022metagait}, outperform the 2D convolution-based methods, like GaitSet \cite{chao2019gaitset} and GaitPart \cite{fan2020gaitpart}, in terms of accuracy, but at the cost of a significant increase in FLOPs (floating point operations). Our approach has a better trade-off between accuracy and efficiency.  The main reason is that the proposed hierarchical learning architecture can extract multi-level motion features while reducing the number of 3D convolutions. More experimental results and ablation analysis are provided in the \textsl{supplementary material}.

\section{Conclusion}
\label{sec:conclu}
This paper presents a hierarchical spatio-temporal representation learning (HSTL) framework for gait recognition. HSTL stacks multiple adaptive region-based motion extractors (ARMEs) and learns walking patterns in a coarse-to-fine manner. An adaptive spatio-temporal pooling (ASTP) module is proposed to perform hierarchical feature mapping for the output of each level of ARME. Additionally, a frame-level temporal aggregation module (FTA) is designed to compress local clips by fusing temporal information. The effectiveness of the proposed HSTL framework is demonstrated through extensive experiments conducted on four public datasets (CASIA-B, OUMVLP, GREW, and Gait3D).

\section*{Acknowledgements}
This work was supported by the National Natural Science Foundation of China (Grant Nos. 61972132, 62106065) and the Research Project for Self-cultivating Talents at Hebei Agricultural University (Grant No. PY201810).



\clearpage
{\small
\bibliographystyle{ieee_fullname}
\bibliography{hstl_final_paper}
}

\end{document}